\documentclass[twoside,11pt]{article}

\usepackage{jmlr2e}
\usepackage{float}
\usepackage{amsmath,amssymb,nccmath}
\usepackage{graphicx}
\usepackage{subfig}
\usepackage{wrapfig}
\usepackage{tabularx}
\usepackage{makecell}
\usepackage{cellspace}
\usepackage[skip=1ex]{caption}

\usepackage[toc,title,page]{appendix}

\usepackage{enumitem}
\setitemize{itemsep=-2pt}
\makeatletter
\newsavebox\myboxA
\newsavebox\myboxB
\newlength\mylenA

\newcommand*\xoverline[2][0.75]{%
    \sbox{\myboxA}{$\m@th#2$}%
    \setbox\myboxB\null
    \ht\myboxB=\ht\myboxA%
    \dp\myboxB=\dp\myboxA%
    \wd\myboxB=#1\wd\myboxA
    \sbox\myboxB{$\m@th\overline{\copy\myboxB}$}
    \setlength\mylenA{\the\wd\myboxA}
    \addtolength\mylenA{-\the\wd\myboxB}%
    \ifdim\wd\myboxB<\wd\myboxA%
       \rlap{\hskip 0.5\mylenA\usebox\myboxB}{\usebox\myboxA}%
    \else
        \hskip -0.5\mylenA\rlap{\usebox\myboxA}{\hskip 0.5\mylenA\usebox\myboxB}%
    \fi}
\makeatother

\ShortHeadings{Neuron Specific Dropout}{Joshua Shunk}
\firstpageno{1}

\begin{document}
\title{Neuron-Specific Dropout: A Deterministic Regularization Technique to Prevent Neural Networks from Overfitting \& Reduce Dependence on Large Training Samples}

\author{\name Joshua Shunk \email joshuashunk@gmail.com \\
       \addr
       Perry High School\\
       1919 E Queen Creek Rd\\
       Gilbert, AZ 85297-0329, USA}

\maketitle

\begin{abstract}%
In order to develop complex relationships between their inputs and outputs, deep neural networks train and adjust large number of parameters. To make these networks work at high accuracy, vast amounts of data are needed. Sometimes, however, the quantity of data needed is not present or obtainable for training. Neuron-specific dropout (NSDropout) is a tool to address this problem. NSDropout looks at both the training pass, and validation pass, of a layer in a model. By comparing the average values produced by each neuron for each class in a data set, the network is able to drop targeted units. The layer is able to predict what features, or noise, the model is looking at during testing that isn't present when looking at samples from validation. Unlike dropout, the ``thinned" networks cannot be ``unthinned" for testing. Neuron-specific dropout has proved to achieve similar, if not better, testing accuracy with far less data than traditional methods including dropout and other regularization methods. Experimentation has shown that neuron-specific dropout reduces the chance of a network overfitting and reduces the need for large training samples on supervised learning tasks in image recognition, all while producing best-in-class results.
\end{abstract}%

\begin{keywords}
  neural networks, regularization, model combination, deep learning, dropout
\end{keywords}

\section{Introduction}

Deep neural networks can understand complex relationships between their inputs and their outputs. By utilizing thousands, or even millions, of hidden nodes (neurons), these models can generate their own set of rules good enough to predict cancer or drive a car. To do this, however, mass amounts of data are needed to train, and then validate, the model. When a non-sufficient amount of data is present, models can become focused on imperfection or sampling noise in the training data. In other words, the model will find details present in the training data that may not be present in its real-world applications. These ultimately lead to overfitting, and because there is no way to make a ``perfect" set of data, other methods have been developed to try to reduce a model's tendency to overfit. One of the most popular methods is to stop the training when the model's validation accuracy starts to diverge from its training accuracy. Another is to implement weight penalties such as L1 and L2 and soft weight sharing.  \\

\setlength{\parindent}{15pt} There are now a couple of ways we can look at solving the problem of overfitting. One is the use of the Bayesian method. Bayesian models work by constructing statistical models based on Bayes’ Theorem.

\[p(\theta | x) = \frac{p(x|\theta)p(\theta)}{p(x)}\]

The goal of a Bayesian ML model is to estimate the posterior distribution ($p(\theta|x)$) given the prior distribution ($p(\theta)$) and the likely hood ($p(x|\theta)$). What makes these models different from classical models is the inclusion of $p(\theta)$ or the prior distribution. One popular prior distribution is the Gaussian process. By taking the average of all parameter settings and weighting its value against the posterior probability given the training data. With the prior Gaussian distribution we can assume that the posterior distribution is normal or falls on a normal bell curve. Assuming we had unlimited computing power, the best way to prevent overfitting would be to calculate a perfect posterior distribution. Approximating the posterior distribution, however, has been proven to provide great results on small models. 

For models with small amounts of hidden nodes, averaging the values of different models trained using different architectures and data can improve the performance compared to a single model. For larger models, however, this process would be too resource-intensive to justify the returns. Training multiple models is difficult because finding optimal parameters can be time-intensive and training multiple large networks can use a lot of resources. In addition, acquiring enough data to train multiple networks, all on different subsets of data is infeasible. Finally, assuming you were able to train multiple networks of different architectures, with different subsets of data, using all of these models when it comes time to test would take too much time in applications that require fast processing. 

\begin{figure}[!tbp]
  \centering
  \captionsetup{format=hang}
  \subfloat[Standard Neural Network.]{\includegraphics[width=0.4\textwidth]{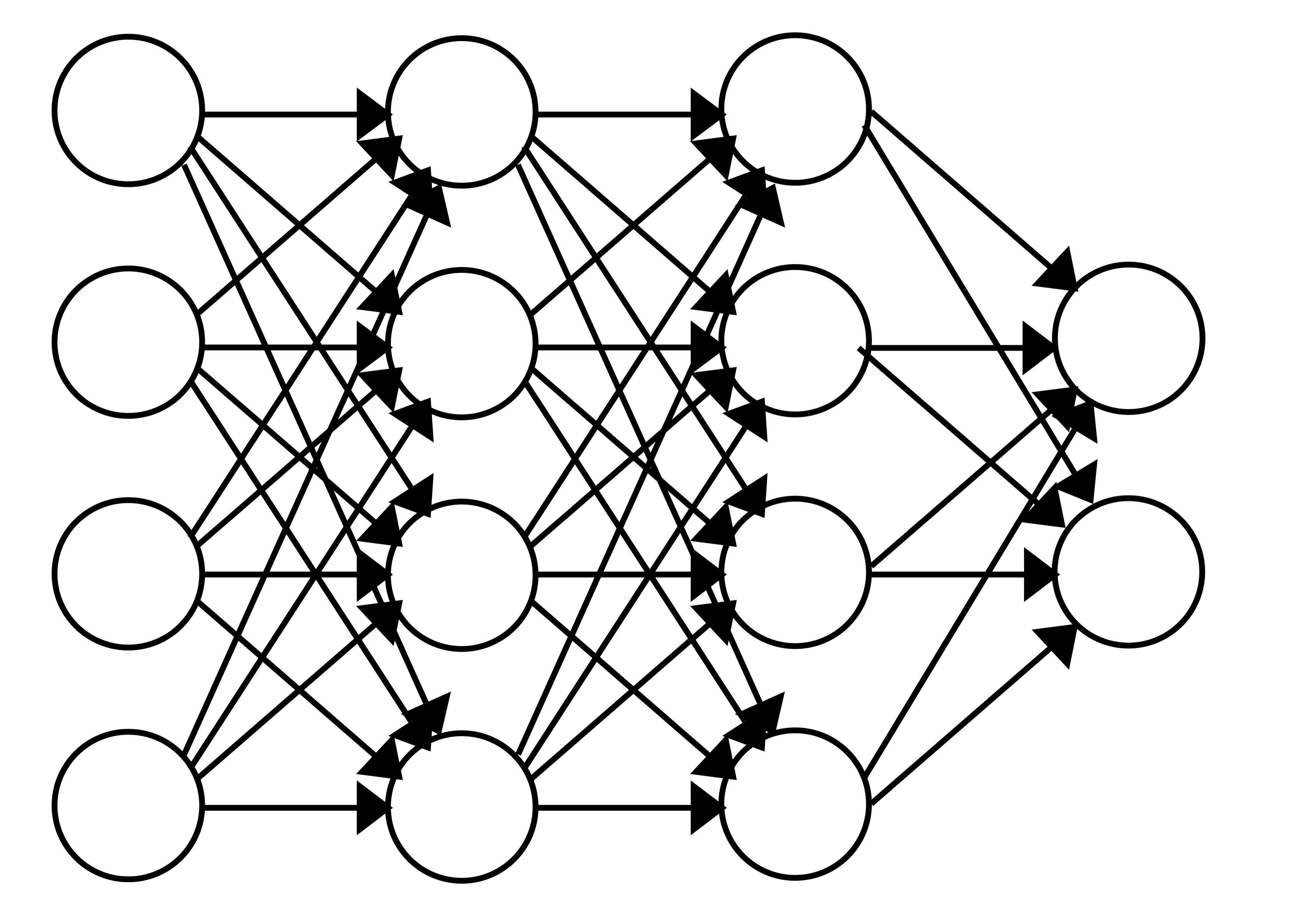}\label{fig:f1}}
  \hspace{5mm}
  \subfloat[After applying NSDropout.]{\includegraphics[width=0.4\textwidth]{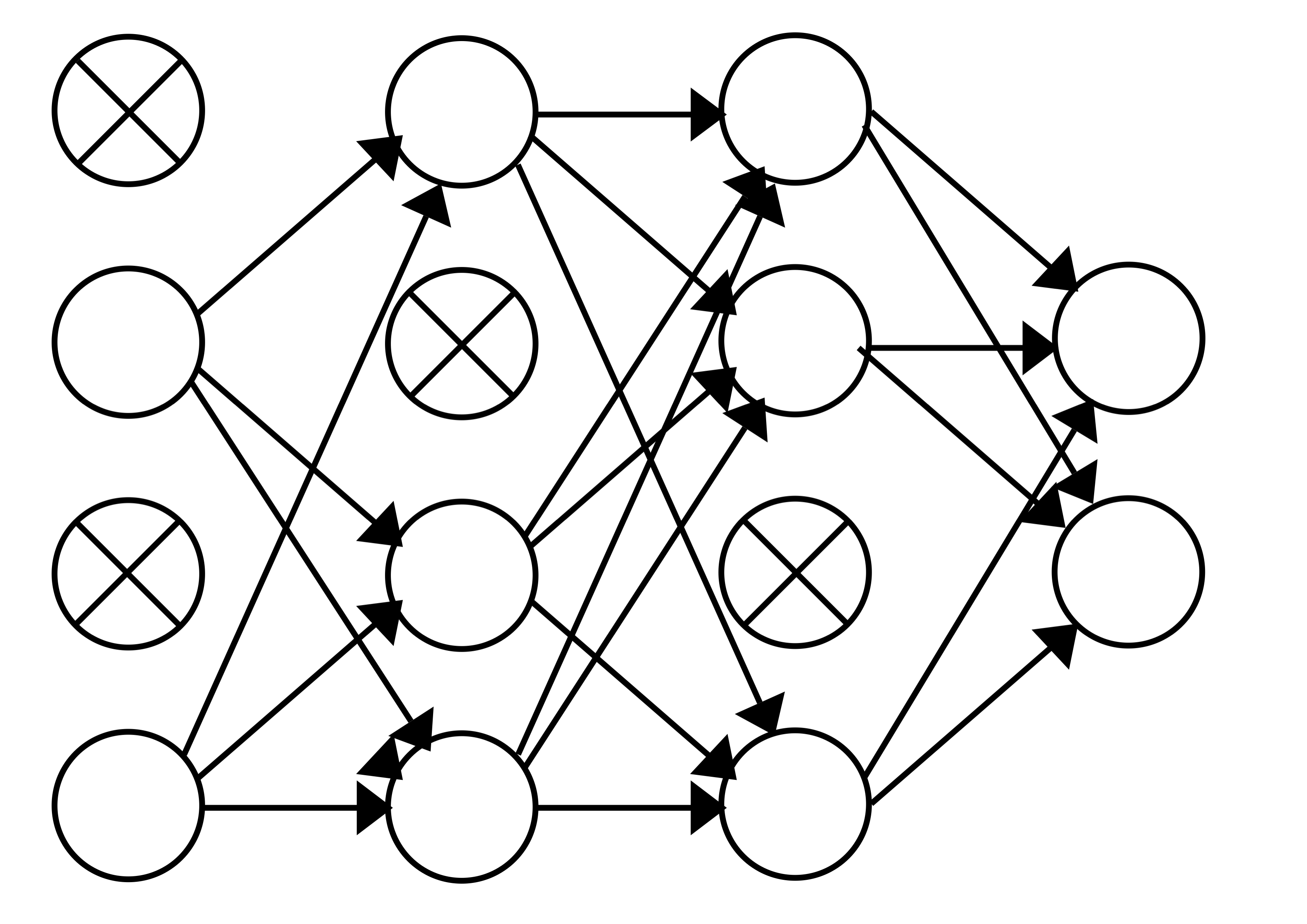}\label{fig:f2}}
  \caption{Neuron Specific Dropout Neural Network Model. \textbf{Left:} A standard neural network with 2 hidden layers, each with 4 hidden units. \textbf{Right:} An example of the network on the left thinned by applying NSDropout.}
  \vspace{-1em}
\end{figure}

This leads to our second option to prevent overfitting. Dropout is a simple yet effective way to limit the influence of noise on a model. It prevents models from overfitting by ``dropping" hidden or visible units, essentially training multiple models simultaneously. By dropping a unit, the unit has no more influence on a model and its decisions during that step. The number of neurons dropped is determined by probability, $p$, that is independent of other units. 

Now we have an additional way to prevent overfitting. Neuron-specific dropout takes the idea of dropping hidden, or visible, units from a layer but doesn't drop them randomly. Unlike other popular layers, neuron-specific dropout takes in four inputs: layer input, the true value of the sample, validation layer input, and the true value of the validation sample. By understanding which neurons have values farthest from the validation average for that class of sample, we can find where noise, or artifacts from training data, are influencing our models' decisions. The number of neurons dropped is determined by the proportion, $p$. This is unlike dropout however, as the probability, $p$, represents what percent of the units in a layer will be dropped. For example, if $p$ is set to 0.2 in a layer with 20 units, there will always be four units dropped. 

Usually, validation data used in a neural network should never influence a model's behavior outside of tuning hyperparameters, but neuron-specific dropout provides an increase in accuracy such that a traditional training data set can be split so that the reserved validation data is never used. A split of training data, so that 20\% is reserved for the new validation set, seems to be optimal.

Similar to dropout, applying neuron-specific dropout produces a ``thinned" neural network. The thinned neural network holds all values from neurons that survived the dropout of neurons. While it could be interpreted that a neural network with $n$ units represents $2^n$ possible thinned neural networks, it is known that the number of different units that get dropped from one step to the next decreases as training progresses. Likewise, the total number of trainable parameters is still $O(n^2)$, or less.

\begin{figure}[!tbp]
    \centering
    \captionsetup{format=hang}
    \subfloat[At training time]{\includegraphics[width=0.4\textwidth]{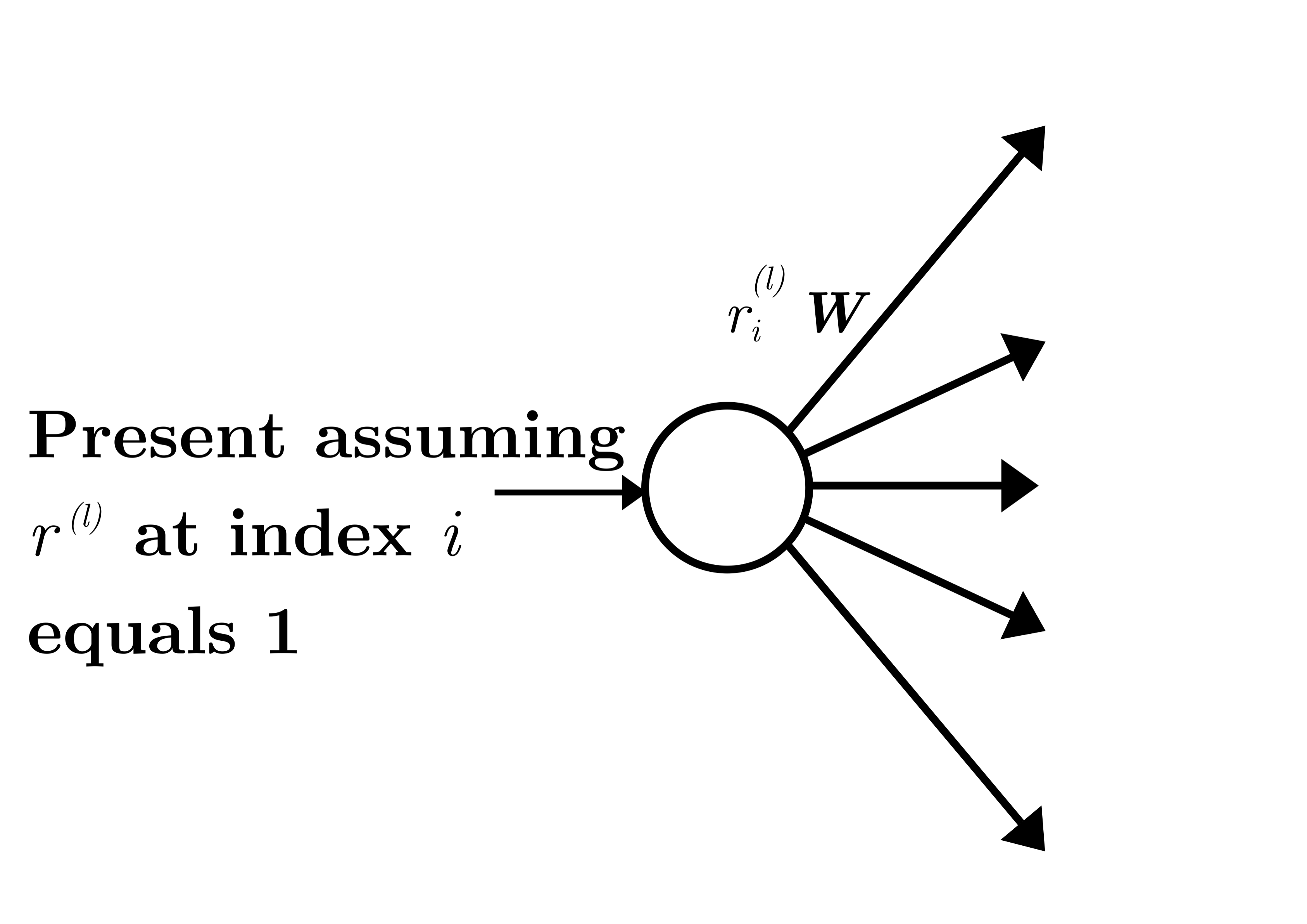}\label{fig2:f1}}
    \hspace{5mm}
    \subfloat[At test time]{\includegraphics[width=0.4\textwidth]{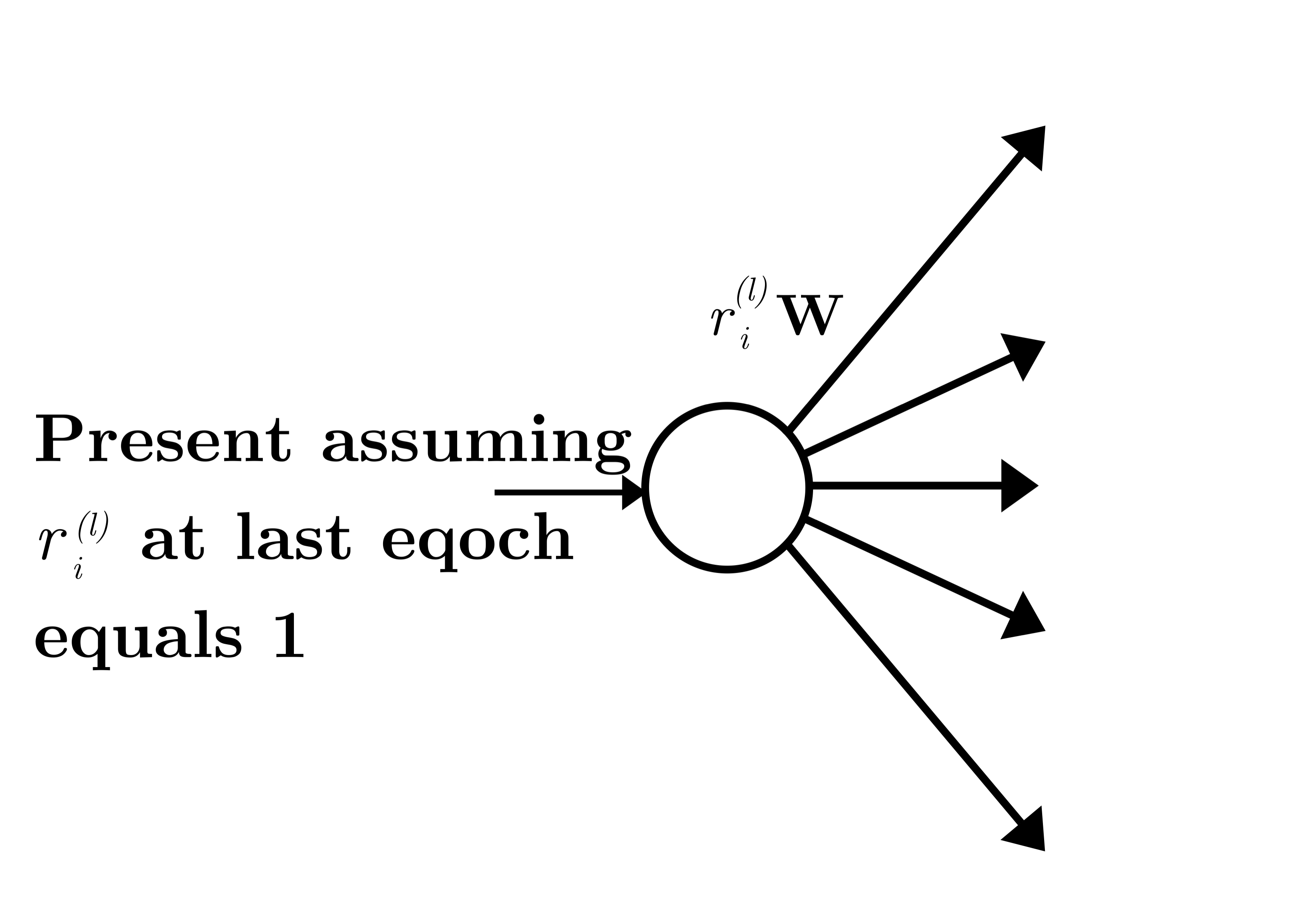}\label{fig2:f2}}
    \caption{\textbf{Left:} A unit at training time that is present assuming the value $r^{(l)}$ at index $i$ is 1. The value of $r_i^{(l)}$ is 1 assuming the output of function $a_i^{(l)}$ is not in the lowest $p$ percentage of values for the output of vector function $a^{(l)}$ . \textbf{Right:} At test time, the unit is present only if the final value of $r_i^{(l)}$ is 1.} 
    \vspace{-1cm}
\end{figure}

Unlike dropout, the benefits of using the layer are not shown if a single, scaled-down, neural net is used. The best results in testing were found when the last used mask was applied. This makes sense because, unlike dropout, units are not dropped randomly. As the model begins to find the units that are influenced by noise and features, it will zero them out, and bringing them back would bring back the weights it has learned to improve without.

This paper is structured as follows. Section 2 describes the motivation for neuron-specific dropout. Section 3 describes relevant previous work. Section 4 formally describes the neuron-specific dropout model and how it works. Section 5 gives an algorithm for training neuron-specific dropout networks and introduces the idea of unseen validation. Section 6 presents experimental results where NSDropout is applied and compares it with other forms of regularization and model combination. Section 7 discusses NSDropout's salient feature and analyzes the effect of neuron-specific and how different parameters can alter the performance of the network.

\section{Motivation}

A motivation for neuron-specific dropout comes from dropout~\citep{JMLR:v15:srivastava14a}. Similar to neuron-specific dropout, dropout shuts off connections to neurons. Originally started as an idea to limit the amount of data, the research quickly shifted when it was discovered that neuron-specific dropout can also help reduce overfitting. In day-to-day life, people learn more information than needed, whether that be from conversations, news, or lessons. Some of the learned information naturally becomes lost when the brain determines that the information wouldn't likely be used in the future. This helps to keep the brain from becoming cluttered. 

One possible explanation for why this occurs is a phenomenon in the brain known as interference. Interference occurs when a memory interferes with other memories. Memory can be defined as a procedure in the brain to acquire, store, retain, and later retrieve information. Interference can be either proactive or retroactive. Proactive interference refers to the brain's inability to remember information because of an older memory. Retroactive interference refers to the brain's ability to retain previously learned information when presented with new information. Neuron-specific dropout uses a method similar to retroactive interference. While models can't inherently know what information will be useful (similar to the human brain), validation data can give them an idea of what they will see when testing. By knowing what noise is present during validation, models can shut off, or forget, what information is not necessary for testing. When each hidden unit is presented with new information -- the outputs of the previous layers -- it takes in and ``learns" from that information, then, before activation, it decides which information ``interferes" with the information presented from validation data. 

\section{Related Work}

Neuron-specific dropout can be seen as an extension to the work done to create dropout~\citep{JMLR:v15:srivastava14a}. Regularization techniques including dropout, L1, and L2 make slight modifications to the algorithms in a way that presents overfitting. Neuron-specific dropout can be interpreted as an extension to these techniques. The idea of adding noise has been further studied by ~\cite{10.1145/1390156.1390294} in the context of Denoising Autoencoders. A model was given the task of denoising the input units of an autoencoder where noise was added. While similar to dropout in the sense that neuron-specific dropout extends this work to show that it can be integrated efficiently into hidden layers, it can not be interpreted as a form of model averaging.

Scaling down the weights of a dropout network by multiplying it by the probability of appearance $p$ at testing allows the model to average the exponential number of thinned networks made during training. The probability of appearance of any given hidden unit in a neuron-specific neural network is not given by the random probability $p$. Instead, $p$ in a neuron-specific neural network represents the proportion of neurons in any given layer $l$ that get ``dropped." Methodology for when a neuron gets dropped can be found in detail in the model description.

Neuron-specific dropout can be seen as a \textit{deterministic} regularization technique compared to dropout which can be seen as \textit{stochastic}. In 2013, ~\cite{pmlr-v28-vandermaaten13} explored similar deterministic regularizers that corresponded to different exponential-family noise distributions. In his work, he explored the idea of dropout which he referred to as ``blankout noise". ~\cite{pmlr-v28-wang13a} explored speeding up dropout by marginalizing its noise and ~\cite{chen2012marginalized} explored marginalization as it relates to denoising autoencoders. 

In work done by ~\cite{nightmareattest} and ~\cite{missingandcorruptedfeatures} explored the idea of dropped with a fixed number of units dropped, and idea present in NSDropout. ~\cite{10.1007/s10994-015-5486-z} explored the idea of using validation results to influence networks when exploring a Bayesian model on multiple data sets. \cite{lengerich2021dropout} explored dropout as it relates to high-order interactions.

\section{Model Description}

This section describes how neuron-specific dropout works how and how it is applied mathematically in a neural network. To do this we are going to assume that images $x_{1}, x_{2}, x_{3}, \cdots, x_{N}$ are passed through the neural network with \textit{L} hidden layers. Let $\textit{l} \in \{1, \cdots, \textit{L}\}$ index the hidden layers of our neural network. A feed-forward operation of the standard network (for $\textit{l} \in \{1, \cdots, \textit{L-1}\}$)  can be modeled by the equation 

\[z_i^{(l+1)} = w_i^{(l+1)}y^l + b_i^{(l+1)}\]
\[y_i^{(l+1)} = f(z_i^{(l+1)})\]

\noindent
where $z_i^{(l+1)}$ represents the input vector of the given layer {l}, $y_i^{(l+1)}$ represents the output vector from layer \textit{l} (Assume the layer given by $y^{(0)}$ represents the networks input). $W^{(l)}$ and $b^{(l)}$ represent the weights and biases of the hidden unit \textit{i}. Let $\textit{i} \in \{1, \cdots, \textit{I}\}$ index the hidden units of our layer \textit{l}. Function \textit{f} is any activation function.\\

The feed-forward operation with neuron-specific dropout is more complicated as you need to take into consideration the whole batch and validation testing. It can be represented by

\begin{align}
a_i^{(l)} &= 
    \begin{pmatrix}
        \begin{bmatrix}
            \begin{bmatrix}
                z_0^{(l+1)} \\           
                \vdots \\
                z_I^{(l+1)}
            \end{bmatrix}_b^c \\
            \vdots \\
            \begin{bmatrix}
            z_0^{(l+1)} \\           
            \vdots \\
            z_I^{(l+1)}
            \end{bmatrix}_b^c
        \end{bmatrix},
        \begin{bmatrix}
            t_0 \\           
            \vdots \\
            t_B
        \end{bmatrix}
        \xrightarrow{argsort}
        \begin{bmatrix}
            \begin{bmatrix}
                z_0^{(l+1)} \\           
                \vdots \\
                z_I^{(l+1)}
            \end{bmatrix}_b^0 \\
            \vdots \\
            \begin{bmatrix}
            z_0^{(l+1)} \\           
            \vdots \\
            z_I^{(l+1)}
            \end{bmatrix}_b^C
        \end{bmatrix},
     \end{pmatrix}
\end{align}
\\

\[s_t^{(l)} = f(a_i^{(l)})\]
\[s_v^{(l)} = f(a_i^{(l)})\]
where $t_b$ denotes the true class value of the image passed through layer \textit{l}. Let $\textit{b} \in \{1, \cdots, \textit{B}\}$ index the samples in the batch. $s^{(l)}$ is the sorted input vector array (\textit{t} and \textit{v} are the testing and validation inputs). Let $\textit{c} \in \{1, \cdots, \textit{C}\}$ index the true label of sample \textit{b}. The goal of function \textit{s} is to sort the inputted z. Let $a_i^{(l)}$ represent the function used by $s_t$ and $s_v$ to sorted the inputted vector by acceding class. Function $a_i^{(l)}$ takes $z_i^{(l+1)}$ and the true class values $t^b$ and returns $z_l^{(l)}$ put in order by class. Now we have the following vector sorted by class:
\begin{align}
        \begin{bmatrix}
            \begin{pmatrix}
                \begin{bmatrix}
                    z_0^{(l+1)} \\           
                    \vdots \\
                    z_I^{(l+1)}
                \end{bmatrix}_b^0 \\
                \vdots \\
                \begin{bmatrix}
                    z_0^{(l+1)} \\           
                    \vdots \\
                    z_I^{(l+1)}
                \end{bmatrix}_b^0 
            \end{pmatrix},
            \hdots, 
            \begin{pmatrix}
                \begin{bmatrix}
                    z_0^{(l+1)} \\           
                    \vdots \\
                    z_I^{(l+1)}
                \end{bmatrix}_b^C \\
                \vdots \\
                \begin{bmatrix}
                    z_0^{(l+1)} \\           
                    \vdots \\
                    z_I^{(l+1)}
                \end{bmatrix}_b^C 
            \end{pmatrix}\\
        \end{bmatrix}
\end{align}

To understand what happens next lets take a close look at one class \textit{c} represented in our network. We will represent the inputs of this sample by $z_i^{(l+1)}$. We now need to take the mean value of each sample in class \textit{c} at the hidden unit \textit{i}. For this example let $\textit{b} \in \{1, \cdots, \textit{B}\}$ index the samples in the batch of class c. This can be represented by:

\[\frac{z_{i, 0}^{(l-1)} + z_{i,1}^{(l-1)} + \hdots + z_{i, B}^{(l-1)} }{B}\]

Now we can apply this to all hidden units, \textit{i} and all classes, \textit{c}. The resultant can be interpreted as a 1-dimensional vector of length \textit{C} representing the average input value at hidden unit \textit{i} for class \textit{c}.

\begin{align}
        \begin{bmatrix}
            \begin{pmatrix}
                \begin{bmatrix}
                    \xoverline[1]{z}_0^{(l+1)} \\           
                    \vdots \\
                    \xoverline[1]{z}_I^{(l+1)}
                \end{bmatrix}^0 \\
            \end{pmatrix},
            \hdots, 
            \begin{pmatrix}
                \begin{bmatrix}
                    \xoverline[1]{z}_0^{(l+1)} \\           
                    \vdots \\
                    \xoverline[1]{z}_I^{(l+1)}
                \end{bmatrix}^C \\
            \end{pmatrix}\\
        \end{bmatrix}
\end{align}

To understand what features of an image a model might be focusing on we need to compare the average values to each hidden unit. To do this we just need to take subtract the average value from the output of the layer, $z_i^{(l+1)}$.

\begin{align}
a_i^{(l)} &= 
    \begin{pmatrix}
        \begin{bmatrix}
            \begin{bmatrix}
                z_0^{(l+1)} \\           
                \vdots \\
                z_I^{(l+1)}
            \end{bmatrix}_b^c \\
            \vdots \\
            \begin{bmatrix}
            z_0^{(l+1)} \\           
            \vdots \\
            z_I^{(l+1)}
            \end{bmatrix}_b^c
        \end{bmatrix}
        -
        \begin{bmatrix}
            \begin{bmatrix}
                \xoverline[1]z_c^{(l+1)} \\           
                \vdots \\
                \xoverline[1]z_c^{(l+1)}
            \end{bmatrix}_b^c \\
            \vdots \\
            \begin{bmatrix}
            \xoverline[1]z_0^{(l+1)} \\           
            \vdots \\
            \xoverline[1]z_I^{(l+1)}
            \end{bmatrix}_b^c
        \end{bmatrix}
     \end{pmatrix}
\end{align}

Now that we have a vector of the difference between the average class at the hidden layer and the hidden unit of that class we can check what values are farthest from the testing mean. Once we have found the values farthest from the mean we need to figure out how many neurons we need to turn off by using the product of $\left \lceil{L*p}\right \rfloor$. Finally, we are left with a mask of

\begin{align}
    r_j^{(l)} = 
        \begin{bmatrix}
            \begin{bmatrix}
                z_0^{(l+1)} \\           
                \vdots \\
                z_I^{(l+1)}
            \end{bmatrix}_b^C 
            * 0 || 1\\
            \vdots \\
            \begin{bmatrix}
                z_0^{(l+1)} \\           
                \vdots \\
                z_I^{(l+1)}
            \end{bmatrix}_b^C 
            * 0 || 1\\
        \end{bmatrix}
\end{align}

Here $*$ denotes an element-wise product. For any layer $l$, $\textbf{r}^{(l)}$ is a vector of independent values determined by the function $a_i^{(l)}$ for each value where the greatest $p$ percentage are multiplied by 0. The resultant vector is multiplied element-wise with $y^{(l)}$, the outputs of layer $l$.  

\[\widetilde{y}^{(l)} = r^{(l)} * y^{(l)}\]

The product is thinned outputs $\widetilde{y}^{(l)}$. The thinned outputs are then used as the input to layer $l+1$, or the next layer. These steps are applied for each layer that has NSDropout creating a feed-forward operation of

\[z_i^{(l+1)} = w_i^{(l+1)}\widetilde{y}^{(l)} + b_i^{(l+1)},\]
\[y_i^{(l+1)} = f(z_i^{(l+1)}).\]

\section{Learning Neuron Specific Dropout Nets}

This section describes a procedure for training neuron-specific dropout neural nets.

\subsection{Backpropagation}

Neuron-specific dropout neural networks can be trained using stochastic gradient descent in a manner similar to normal dropout networks and standard neural networks. For each mini-batch however, a thinned network was sampled by deliberately dropping out select units. Forward and backpropagation are done on this thinned network, not the full network. As NSDropout is deterministic is is possible for the network to not change from one batch to the next, though unlikely as shown in figure 9. Similar to dropout, any training case which uses no parameter contributes a gradient of zero for that parameter. Common methods to improve stochastic gradient descent, such as L2 weight decay, momentum, and annealed learning rates were found to improve NSDropout neural networks.

Unlike dropout, constraining the norm of the incoming weight vector proved to show no substantial benefits. Dropout sees improvement when the vector of weights inputted, on any hidden unit, is constrained by fixed constant $c$. One possible explanation for this is that weight vectors are already constrained by the validation data as seen in the model description. In this sense, you could consider NSDropout a form of Max-norm regularization. Max-norm regularization has been used before in collaborative filtering ~\citep{10.1007/11503415_37}.

\subsection{Validation Data}

Unlike traditional neural networks that require one set of data for training, neuron-specific neural networks require two. While validation data is often considered part of the training data, because it is used by the research to modify parameters, it never affects how the model learns outside of the parameters. This is the biggest difference between classical neural networks including those that use dropout, and networks, that use NSDropout.

The goal of NSDropout out is not just to find, and correct, a model if it starts to overfit, but to find out where a model is overfitting. In order to do this, the model needs references it hasn't seen during training. This essentially creates four sets of data (training, validation, unseen validation, and testing) compared to the normal three (training, validation, and testing) found in traditional neural networks. Part of NSDropout's goal was to reduce dependence on large data sets. For this, no more data is needed than a traditional network. Training data for NSDropout is split again to create a new validation set. Similar to the popular 80/20 split for training and validation data, it is most effective to split the training data into another 80/20 split with 20 percent being the new validation data. For example, assume a data set of 60,000 images and 10,000 testing images. A traditional network would see 48,000 training and 12,000 validation images. A NSDropout model would see 38400 training images, 96,000 validation images, and 12,000 unseen validation images. The 10,000 testing images would remain the same to be used for evaluation at the end of training. Less training data can be used to achieve results similar to traditional networks.

During testing, training with as little as 15\% of the original data (including new validation data) was proven to improve the testing accuracy of the network. A possible explanation is that NSDropout drops the noise found from such small data sets. More information on the effect of data set size on accuracy can be found in section 7.2.

\section{Experimental Results}
Neuron-specific dropout was tested on data sets from multiple different domains for classification. It was found that neuron-specific dropout improved generalization performance on \textit{all} data sets compared to neural networks that used dropout and networks without dropout. A brief description of each data set is given in table 1. The data sets include:
\begin{itemize}
\item MNIST: A standard toy data set of handwritten digits.
\item Fashion-MNIST: A standard toy data set of fashion items.
\item CIFAR-10: Tiny natural images ~\citep{Krizhevsky2009LearningML}.
\end{itemize}

The MNIST and CIFAR-10 data sets were chosen in relation to the original dropout paper and its testing methodology. The data sets were chosen to prove that dropout, as well as neuron-specific dropout, are techniques and can be used on a variety of domains. Located in Appendix B is a more detailed description of all the experiments and data sets.

\begin{center}
\begin{table}[H]
\resizebox{\textwidth}{!}{%
\centering
\begin{tabular}{ l c c c c }
 \hline
 \textbf{Data Set} & \textbf{Domain} & \textbf{Dimensionality} & \textbf{Training Set} & \textbf{Test Set} \\ 
 \hline
 MNIST 								  & Vision 		& 784 (28 × 28 grayscale)      & 10K              & 10K \\
 Fashion-MNIST & Vision & 784 (28 x 28 grayscale) & 10K & 10K\\
 
 CIFAR-10 					 & Vision     & 3072 (32 × 32 color) 			& 25K              & 10K\\ 

 \hline

\end{tabular}}
 \label{table:1}
\caption{Overview of the data sets used in this paper. Training set given for NSDropout.}
\end{table}
\end{center}

\vspace{-3em}

\subsection{Results on Image Data Sets}

Three different data sets were used to evaluate NSDropout -- MNIST Handwritten Digits (Numbers), Fashion-MNIST, and CIFAR10. These data sets offer a unique range of image types and training sizes. The models that performed the best \textit{all} utilized NSDropout. 

\subsubsection{MNIST Handwritten Digits}

\hfill 

\begin{center}
\begin{table}[H]
\centering
\begin{tabular}{ l c c c }
\hline
 \textbf{Method} & \textbf{Unit Type}  & \textbf{Architecture} & \textbf{Error \%} \\ 
  \hline
 Dropout NN ~\citep{JMLR:v15:srivastava14a} 								   & ReLU 	& 3 layers, 128 units 			              & 5.59\\
 Standard Neural Net ~\citep{Simard2003BestPF} 		  & Logistic 		& 2 layers, 800 units                    & 1.6 \\
 Dropout NN ~\citep{JMLR:v15:srivastava14a} & ReLU & 3 layers, 1024 units & 1.25 \\
 \thead[l]{Dropout NN + max-norm constraint \\ ~\citep{goodfellow2013maxout}} & \thead{Maxout} & \thead{2 layers, (5 x 240) units} & \thead{0.94} \\
 \hline
DBN + finetuning  &  Logistic  &  500-500-2000  &  1.18 \\
DBM + finetuning  &  Logistic  &  500-500-2000  &  0.96 \\
DBN + dropout finetuning & Logistic & 500-500-2000 & 0.92 \\
DBM + dropoutfinetuning & Logistic & 500-500-2000 & 0.79 \\

\hline

 NSDropout NN 					 & ReLU     & 3 layers, 128 units 			             & \textbf{0}\\

 \hline

\end{tabular}
 \label{table:2}
\caption{Comparison of different models on MNIST Handwritten Digits.}

\end{table}

\end{center}

\vspace{-3em}

The MNIST data set contains 60,000 training and 10,000 testing images. Each image contains a 28 x 28 pixel handwritten digit. The goal is to accurately classify the images into 10 distinct categories. Table 2 compares the performance of NSDropout compared to dropout and a standard neural network. The best performing neural network was a 3 layer, 128 unit NSDropout net with a rectified linear activation unit (ReLU)~\citep{5459469} as activation. More information about model architecture can be found in Appendix B.1. A traditional dropout trained without batch's yielded an error rate of 5.59. Increasing the unit count to 800 and replacing ReLU with a logistic unit, decreases the error to 1.6. Changing the network back to dropout and changing logistic units back to ReLUs and increasing the layer and unit count to 3 and 1024 respectfully decreased the error to 1.25. Using maxout units ~\citep{goodfellow2013maxout}, instead of ReLUs, further reduces the error down to 0.94\%. 

\begin{wrapfigure}{r}{0.5\textwidth}
  \captionsetup{format=hang}
  \vspace{-0.8cm}
  \begin{center}
    \includegraphics[width=0.48\textwidth]{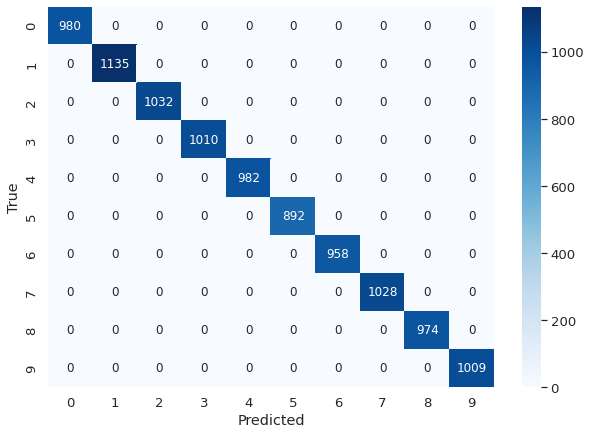}
 \end{center}
  \caption{Confusion Matrix}
\end{wrapfigure}

Using RBMs and Deep Boltzmann Machines further reduced the error shown in Table 2. A pretrained DBM achieved a test error of 0.79\%. Using NSDropout, the test error \% reduces down to 0, a perfect test run. Figure 3 shows a confusion matrix produced by the final model. The model was trained for 359 epochs and hit a perfect unseen validation accuracy 5 times. The model had only a little more than 335,000 parameters, far lower than the millions required to produce error rates under 1 using traditional dropout nets and even DBMs. How NSDropout was able to achieve a perfect test can be found in Appendix B.1.
\begin{figure}[!tbp]
    \centering
    \captionsetup{format=hang}
    \subfloat[MNIST Handwritten Digits ]{\includegraphics[width=0.45\textwidth]{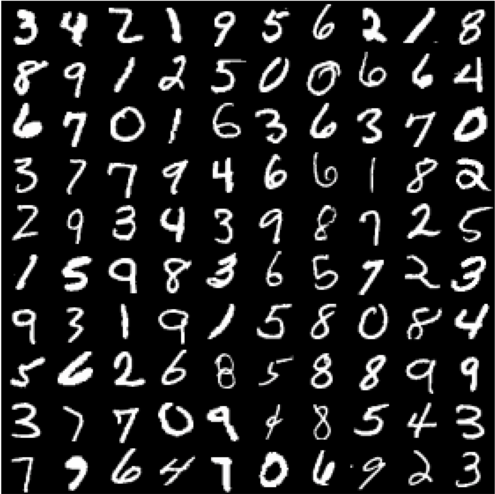}
    \label{fig4:f1}}
    \hspace{5mm}
    \subfloat[Fashion-MNIST]{\includegraphics[width=0.45\textwidth]{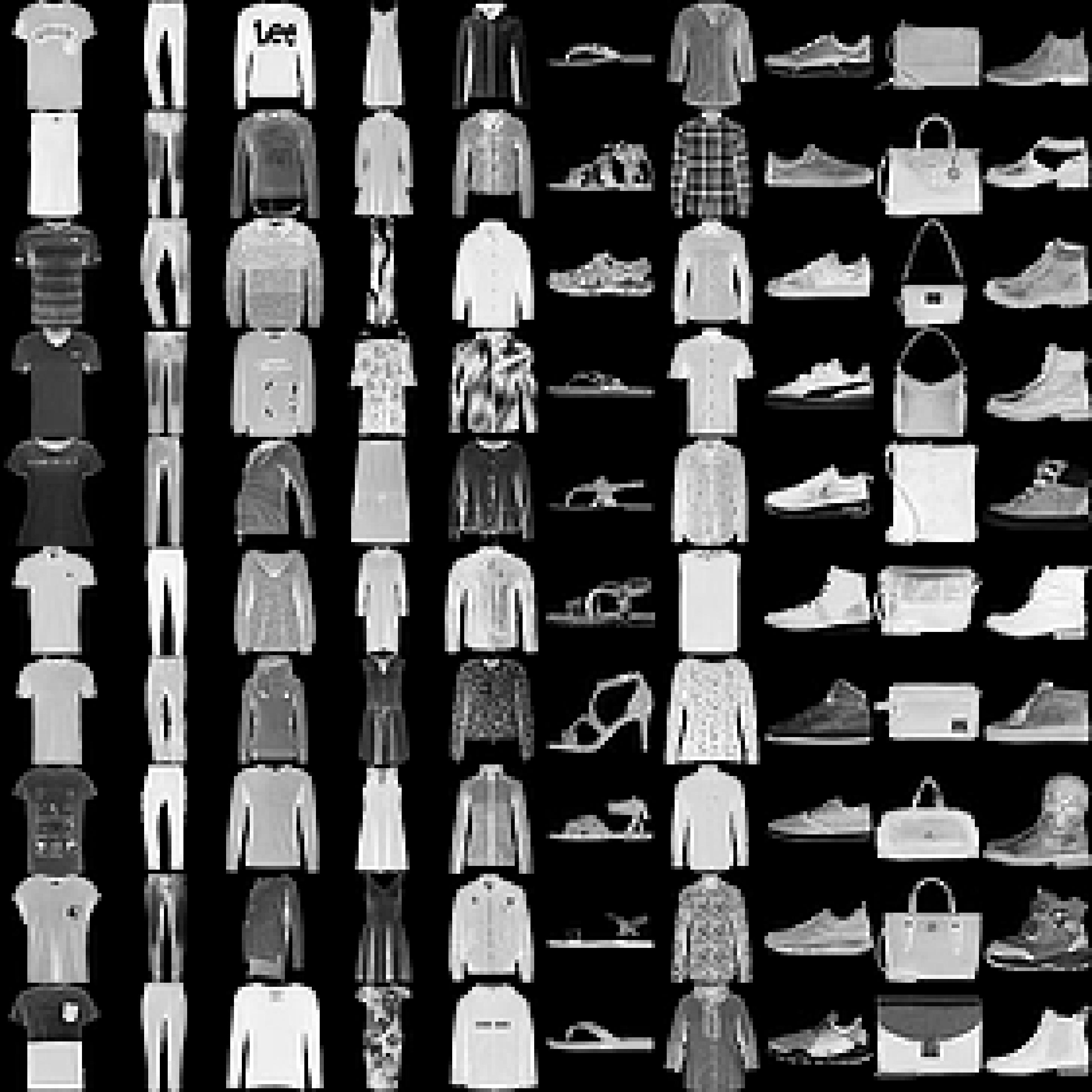}}\label{fig4:f2}
    \caption{Samples from image data sets. Columns in figure b represent different categories.} 
    \vspace*{-\baselineskip}

\end{figure}
\subsubsection{Fashion-MNIST}

\vspace{-10pt}
\begin{center}
\begin{table}[H]
\centering
\begin{tabular}{ l c c c }
\hline
 \textbf{Method} & \textbf{Unit Type}  & \textbf{Architecture} & \textbf{Error \%} \\ 
  \hline
 Standard Neural Net 		  & ReLU 		& 3 layers, 256 units                    & 15.23 \\
 Dropout NN 								   & ReLU 	& 3 layers, 128 units 			              & 12.25\\ 
 NSDropout NN 					 & ReLU     & 3 layers, 128 units 			             & \textbf{0.19}\\ 

 \hline

\end{tabular}
\label{table:3}
\caption{Comparison of different models on Fashion-MNIST.}

\vspace*{-2cm} 
\end{table}

\end{center}
 
\vspace*{-\baselineskip}

Consisting of 60,000 training and 10,000 testing 28 x 28 examples. Each example is a grayscale image, associated with a label from 10 classes. The goal is to accurately classify the images into each of the 10 classes. Table 3 compares the performance of a neuron-specific dropout network to a dropout network and a standard neural network. The best performing neural network was a 3 layer, 128 unit, NSDropout net with a rectified linear activation unit (ReLU)~\citep{5459469} as activation (identical to MNIST Handwritten Digits). NSDropout was applied to the input with a $p$ of 0.2. As noted in Appendix A.3, a $p$ value of 0.2 for an input layer is lower than recommended for most networks but was found to be ideal for Fashion-MNIST.  When testing, Dropout reduced the error \% from 15.23 to 12.25. NSDropout further reduced error to 0.19\%. A 98.75\% decrease from the standard neural network's 15.24\% error.

All models were trained without batches and passed through all training data per epoch. Both NSDropout and Dropout utilized 10K images out of the available 60K for training and validation. The Standard Neural Network utilized all 60K (48,000 for training and 12,000 for validation).

\begin{wrapfigure}{r}{0.5\textwidth}
  \captionsetup{format=hang}
  \vspace{-0.8cm}

  \begin{center}
    \includegraphics[width=0.48\textwidth]{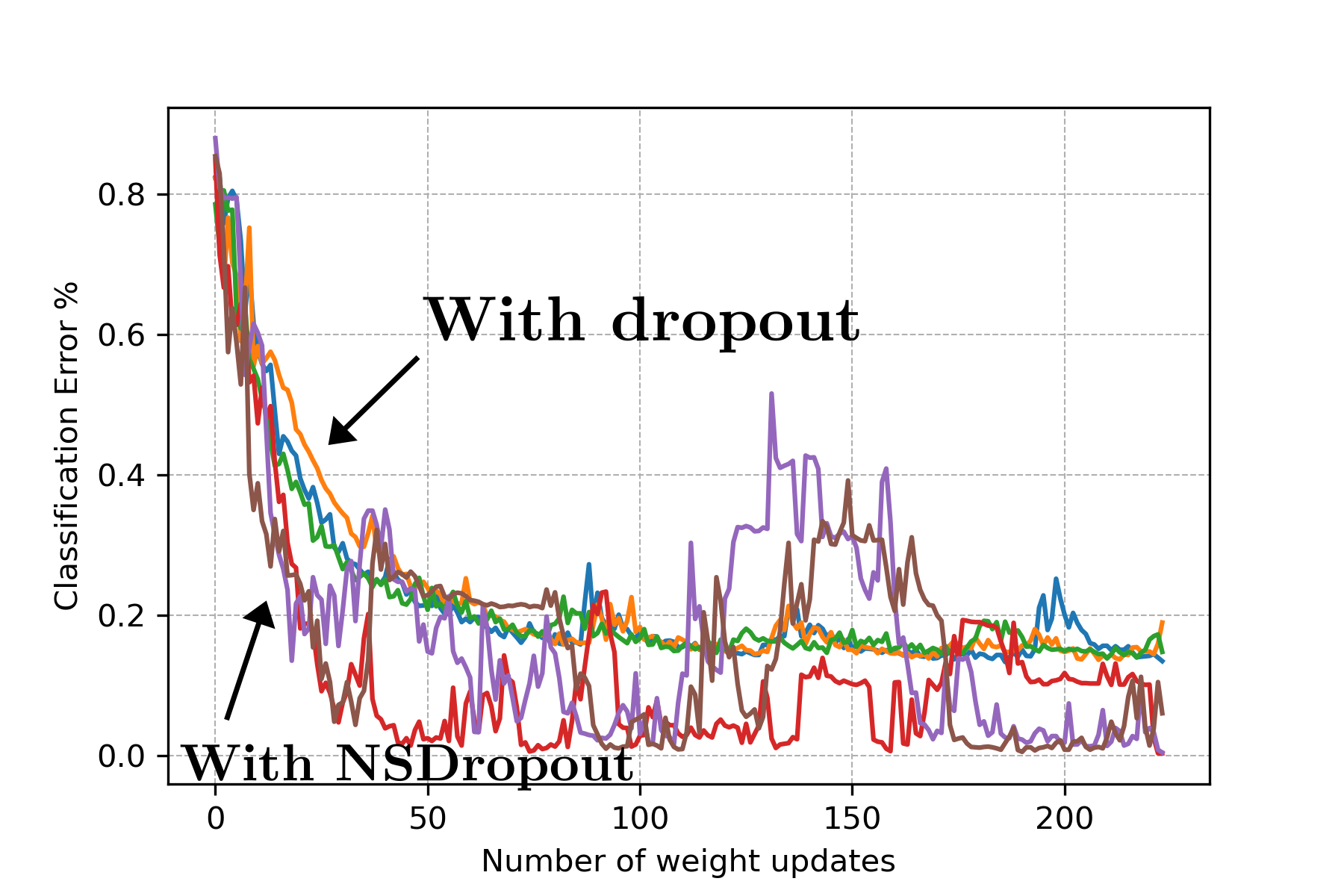}
  \end{center}
  \caption{The error for different runs with dropout and NSDropout. The networks have 3 hidden layers each with 128 units.}
\end{wrapfigure}

Many different parameters and architectures were tested multiple times to verify results. During testing, results would always end up lower than dropout but would fluctuate much more as seen in Figure 3. There are a few possibilities as to why this might occur. A model might slowly build up a dependence on a unit focused on one feature of training images. Once that dependence gets caught by NSDropout it drops it and the model has to relearn without it. Another possibility is the low unit counts of the trained models. 128 hidden units were trained per hidden layer. Increasing the number of hidden units per hidden layer did not increase the accuracy. While it did help limit the sporadic nature, the model started to see lower accuracy at anything above 1024 units. 

\subsubsection{CIFAR-10}
\vspace*{-5pt} 
\begin{center}
\begin{table}[H]
\centering
\begin{tabular}{ l  c }
\hline
 \textbf{Method}  & \textbf{Error \%} \\ 
  \hline
 Dropout Net ~\citep{JMLR:v15:srivastava14a} 	               & 48.78\\ 
 Conv Net + stochastic pooling ~\citep{zeiler2013stochastic} & 15.13\\ 
 Conv Net + max pooling ~\citep{snoek2012stochastic}        & 14.98 \\
 Conv Net + max pooling + dropout fully connected layers ~\citep{JMLR:v15:srivastava14a} & 14.32 \\
 Conv Net + max pooling + dropout in all layers ~\citep{JMLR:v15:srivastava14a}& 12.61 \\
 Conv Net + maxout ~\citep{goodfellow2013maxout} & 11.68 \\
 NSDropout Net                 & \textbf{7.72}\\ 

 \hline

\end{tabular}
 \label{table:4}
\caption{Error rates on CIFAR-10.}
\vspace*{-1cm} 
\vspace*{3pt} 
\end{table}
\end{center}

\vspace*{-\baselineskip}

The CIFAR-10 data set contains 70,000 32 x 32 pixel color images from 10 unique categories. Figure 6 gives a few examples of images from this data set. More information about model architecture, data preprocessing, and a description of the data can be found in Appendix B.3. Table 4 shows the error rate achieved by different methods on the CIFAR-10 data set. Using a 3 layer 256 unit dropout model resulted in an error \% of 48.78. Switching to a convolutional neural network with stochastic pooling decreased error to 15.13. Using max-pooling decreased it further to 14.98. When dropout was reintroduced the error decreases to 12.61. When NSDropout was used, the error rate drops to an impressive 7.72\%. While ZCA whitening ~\citep{BELL19973327} was tested, no substantial benefits were found.
 
\begin{figure}[!tbp]
    \centering
    \captionsetup{format=hang}
    \subfloat[CIFAR-10 ZCA Preprocessing]{\includegraphics[width=0.45\textwidth]{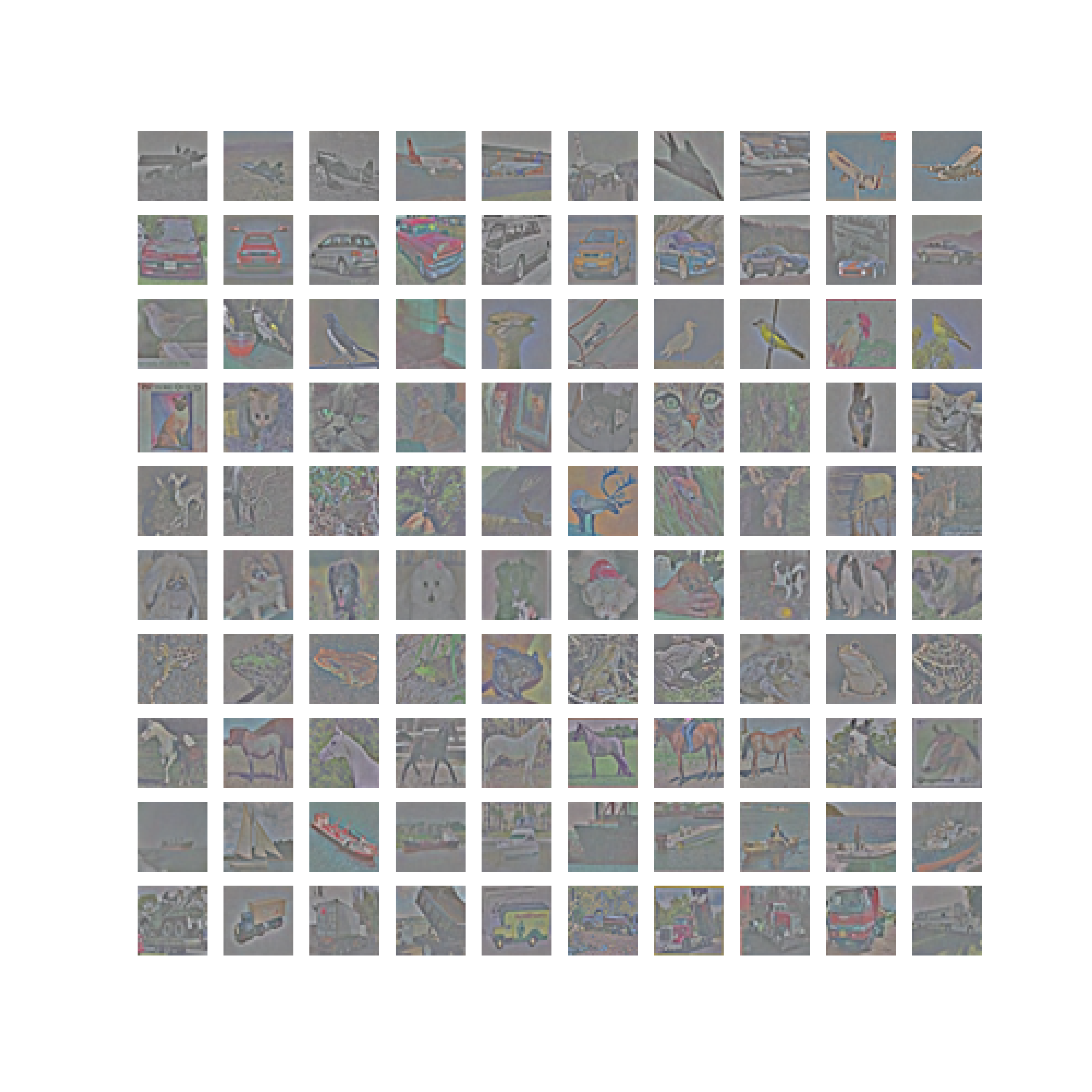}\label{fig6:f1}}
    \hspace{3mm}
    \subfloat[CIFAR-10]{\includegraphics[width=0.45\textwidth]{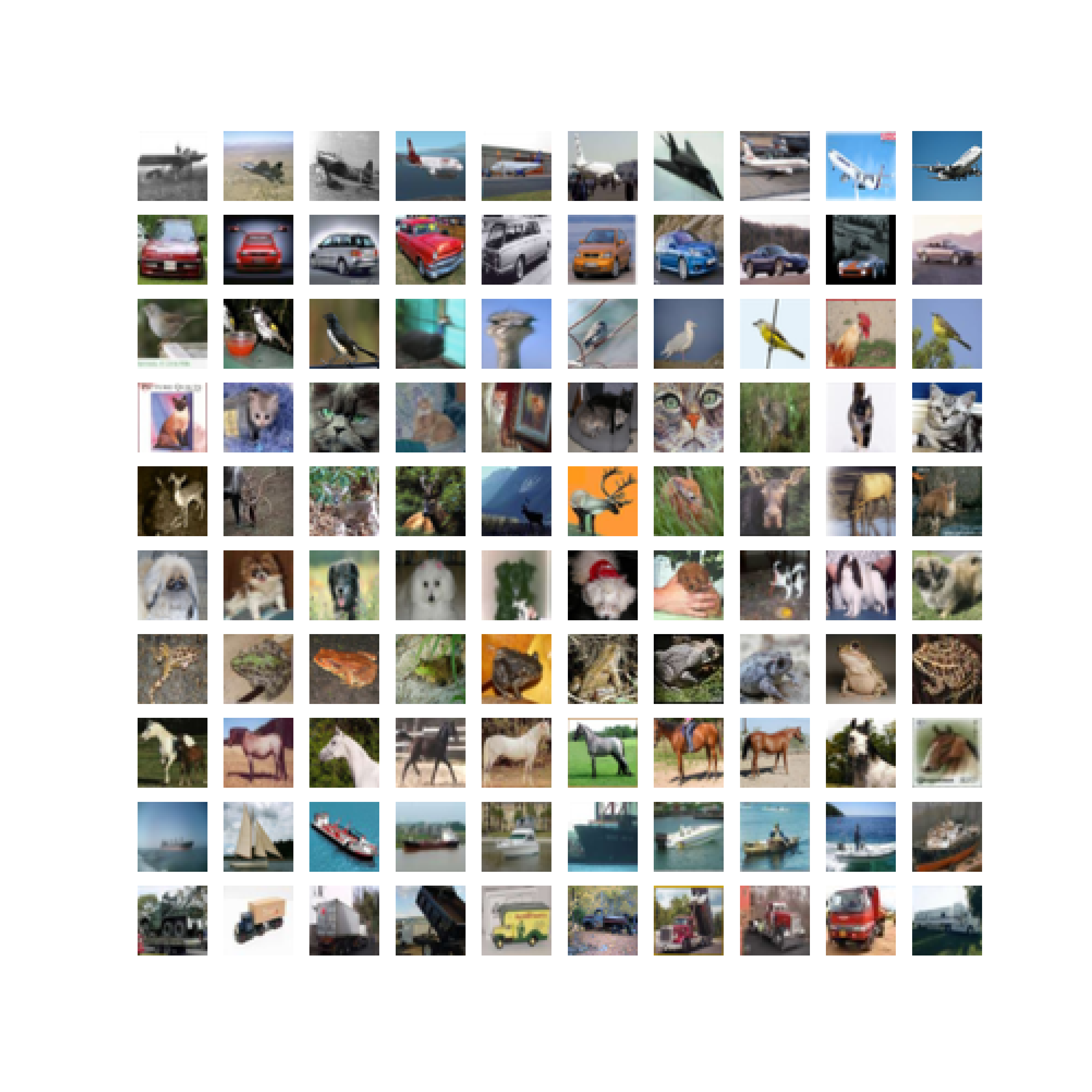}\label{fig6:f2}}
    \caption{Samples from image data sets. Each row corresponds to a different category.} 
\end{figure}

\section{Salient Features}

We know that NSDropout provides improvements over traditional networks and dropout networks from the results of the experiments. In this section, the effects dropout can have on a neural network are closely examined.

\subsection{Effect of Dropout Rate}

NSDropout has one tunable hyperparameter $p$ (the proportion of neurons in layer $l$ that get ``dropped" in the network). In this section, we will explore how changing $p$ affects the error \% and how it compares to changing $p$ in a dropout layer. When comparing testing to training accuracy with varying $p$, the same architecture of 784-128-128-128-10 was used. NSDropout was located after the first activation. Figure 7a shows the obtained test error as a function of $p$. In this scenario $p$ represents the proportion of units in a layer ``dropped". A $p$ value of 0.8 means that only 20\% of units are retained. A $p$ value of 0 means that all units are retained. It can be seen that on average as $p$ increases up to a value of 0.6, the test and train error \% decrease, where it then only marginally increases at 0.7. From this we can see that, within a margin of error, an ideal $p$ value is when 0.3 $\leq p \leq$ 0.7, a slight change from the range of 0.4 $\leq p \leq$ 0.7 proposed by \cite{JMLR:v15:srivastava14a} for keeping $pn$ fixed for dropout. 

\begin{figure}[H]
    \centering
    \captionsetup{format=hang}
    \subfloat[Effect of varying $p$. ]{\includegraphics[width=0.4\textwidth]{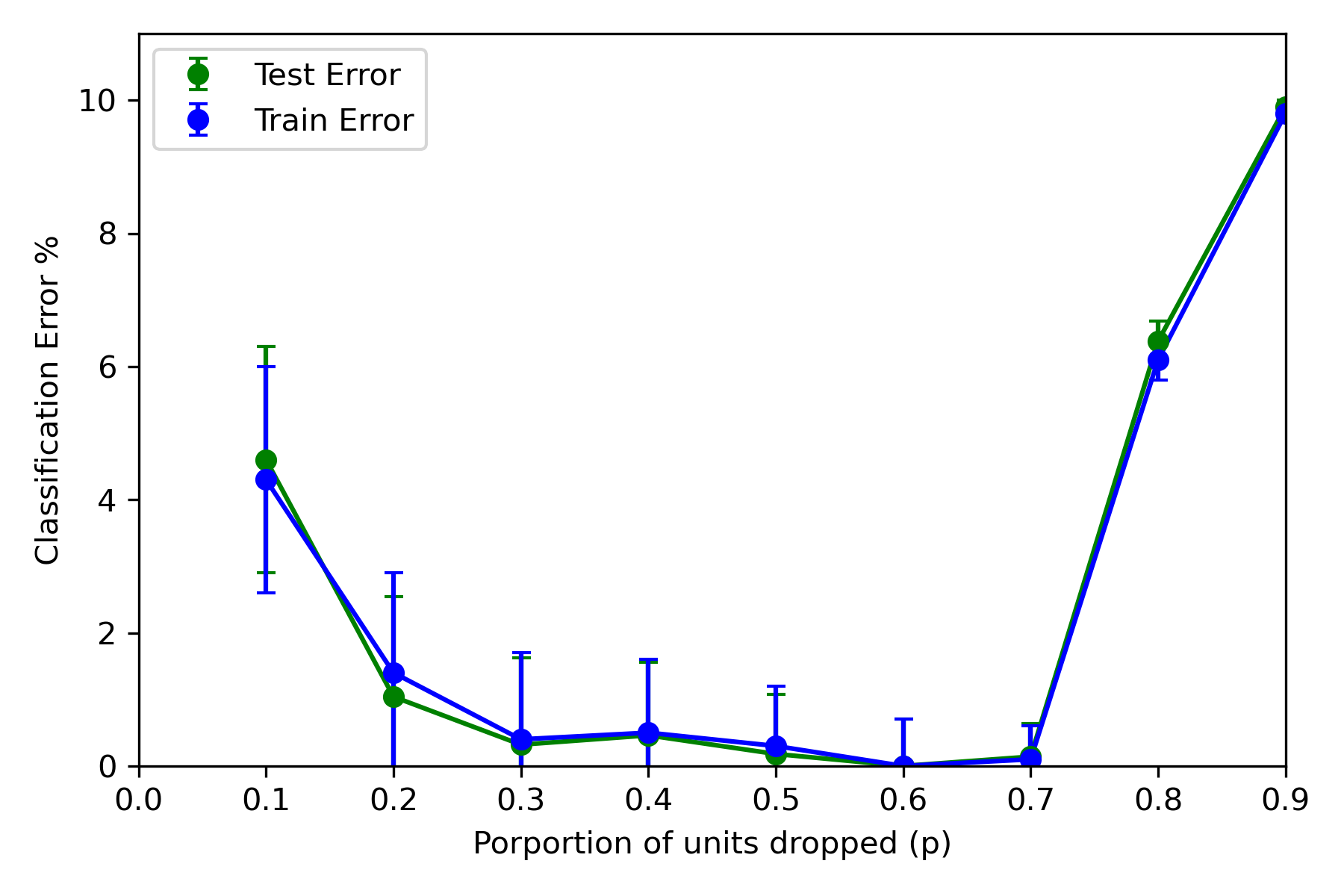}\label{fig7:f1}}
    \hspace{5mm}
    \subfloat[Comparing NSDropout to Dropout. ]{\includegraphics[width=0.4\textwidth]{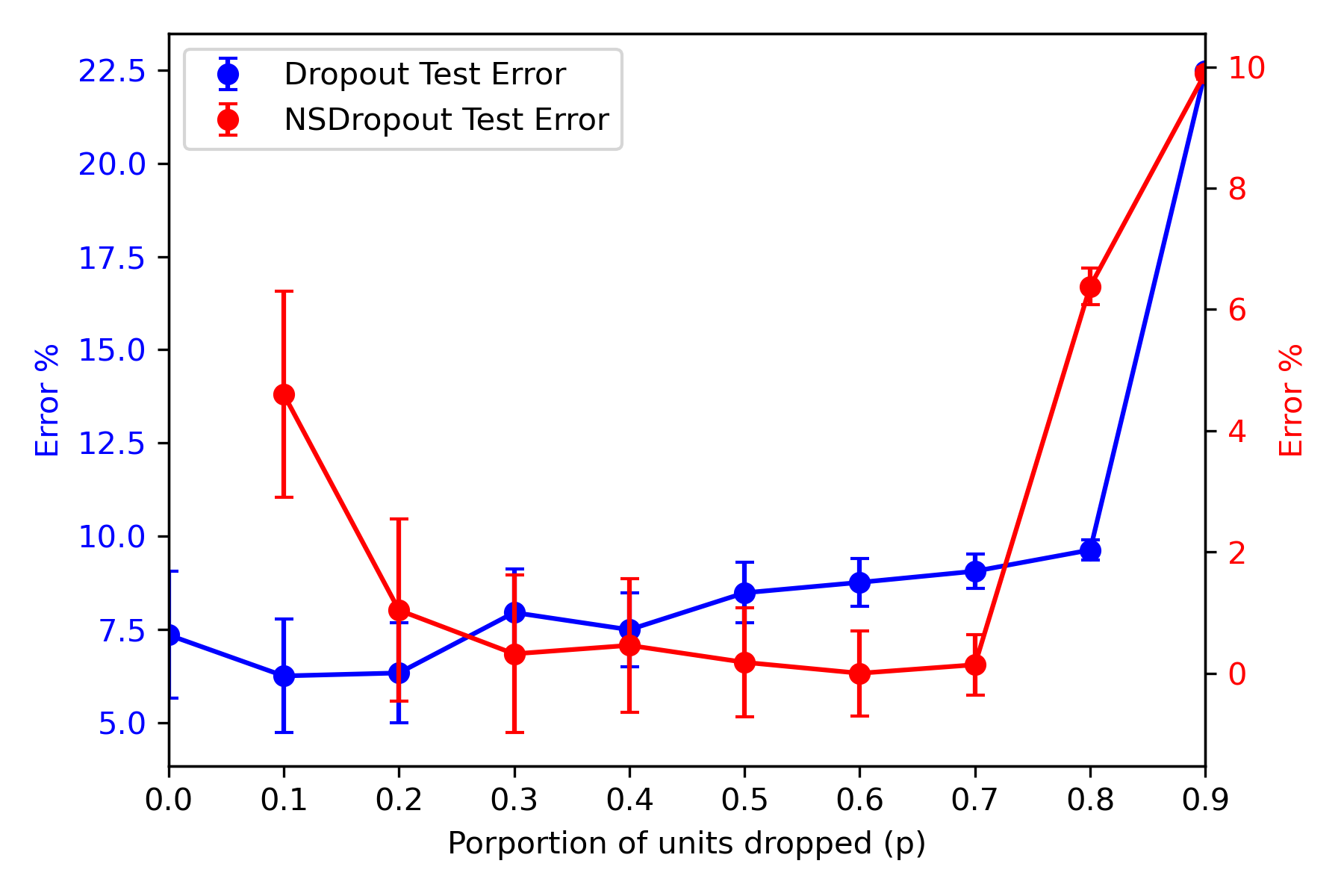}\label{fig7:f2}}
    \caption{\textbf{Left:} Comparing testing error \% to training error \% when changing the value of $p$. \textbf{Right:} Comparing the trends of NSDropout to the trends of Dropout when varying $p$.} 
\end{figure}

When comparing the trends of NSDropout and Dropout in figure 7b we find that for both methods, error \% becomes moderately flat when 0.3 $\leq p \leq$ 0.7. The trends, however, from 0.1 $\leq p \leq$ 0.7 are reversed. As $p$ increases for NSDropout, the error \% decreases while for the same range, error \% for dropout increases. 

The benefit of NSDropout can be seen in figure 7a. The test and train error \% of NSDropout are closely tied together until slight separation when $p = 0.1$. Even when varying $p$ the model shows no sign of overfitting.

\subsection{Effect of Data Set Size}
One of the main goals of NSDropout was to decrease the dependence on large data sets. As a result, this is one of the principle sections of the paper. In a traditional model, decreasing the size of the data set decreases the model's ability to generalize. The opposite is also true, increasing the size of a data set can decrease overfitting and increase its ability to generalize results. A large sample is not always available, however. This section explores the effect of changing the data set size on the error \% of a NSDropout feed-forward network. To see if NSDropout can improve over other models, classification experiments on MNIST Handwritten Digits were conducted with varying amounts of data.

\begin{wrapfigure}{r}{0.5\textwidth}
  \captionsetup{format=hang}
  \vspace{-30pt}
  \begin{center}
    \includegraphics[width=0.48\textwidth]{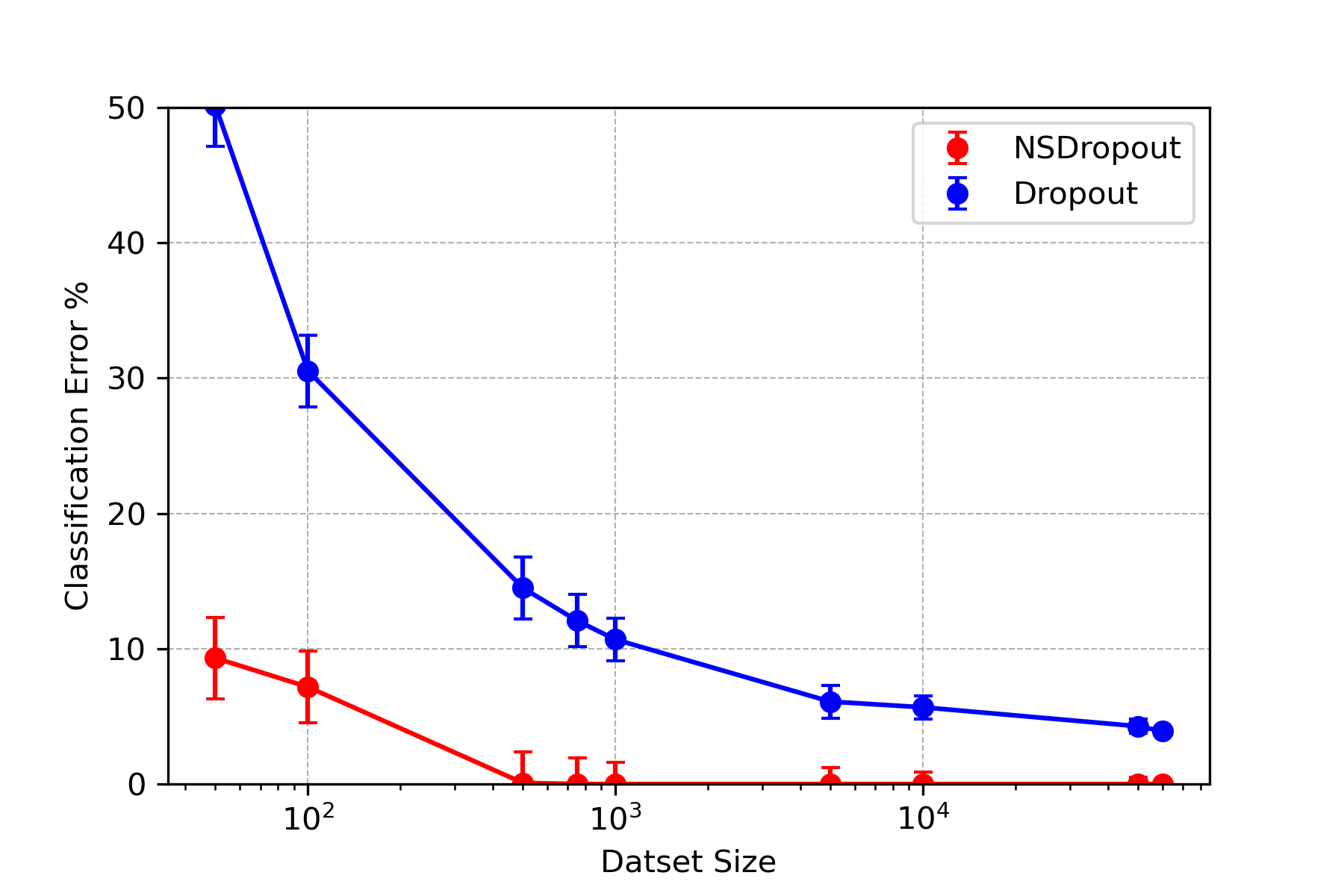}
  \end{center}
  \caption{Effect of data set size on error \%.}
\end{wrapfigure}

The results of these experiments are shown in Figure 7. Data sets of size 60K (all available), 50K, 10K, 5K, 1K, 750, 500, 100, and 50 all randomly chosen, were used for training. The same network architecture (784-128-128-128-10) was used for all data sets for both dropout and NSDropout. Both models used $p = 0.5$ for the input layer. While NSDropout improved over dropout in every experiment, the most notable strength of NSDropout can be found in the data sets of sizes 500 and 750. When given only 500 images (0.83\% of the full data set), NSDropout achieved a near-perfect accuracy of 99.92\%. Its first perfect accuracy came when given only 750 examples. In comparison, dropout had an accuracy of about 88\% when given 750 images. When given a mere 50 images (average of 5 per class), the model was still able to achieve an accuracy of over 90\%. All testing was done on the full 10,000 examples MNIST testing set.

\subsection{Effect on Variability of Mask}

Being deterministic, the question arose how often does the mask created change. If a unit gets ``dropped", causing it to not learn with the rest of the model it was thought that the unit would be stuck in a loop where it would always be the greatest difference between testing and training. To test this, the mask of each figure was tracked from epoch to epoch and compared to see how often the ``dropped" units changed. 

\begin{wrapfigure}{r}{0.5\textwidth}
  \captionsetup{format=hang}
  \vspace{-30pt}
  \begin{center}
    \includegraphics[width=0.48\textwidth]{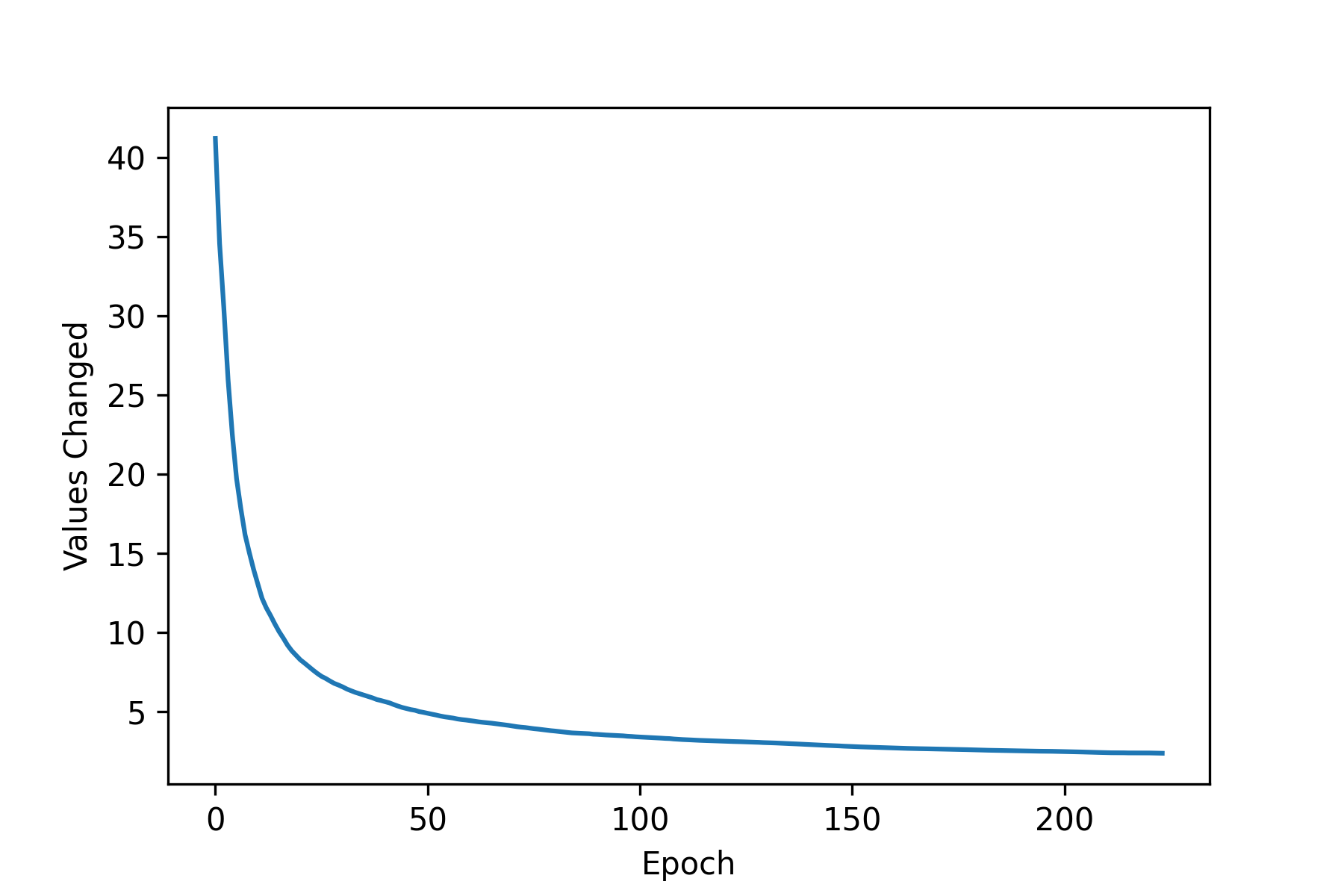}
  \end{center}
  \caption{Mask Values Changed from Epoch to Epoch}
\end{wrapfigure}

Figure 9 is a visual representation of the results of the experiment. Assuming a network is training to recognize $n$ number of classes, NSDropout would produce $n$ number of masks, one for each class. Figure 9 shows on average the number of times a location of a ``dropped" unit changes in each mask from one epoch to the next. The input layer NSDropout was applied to consist of 128 units. From epoch 1 to 2 we can see on average, more than 40 of the 128 units were either ``dropped" for the first time, or were no longer ``dropped" after being dropped in epoch 1.

\section{Conclusion}

Neuron-specific dropout (NSDropout) is a deterministic regularization technique to improve the accuracy of neural networks with an emphasis on networks with small amounts of training data. Through traditional learning techniques, networks develop complex relationships between their inputs and outputs for a set of data. Often however these complex relationships fail to generalize to unseen data. Unlike dropout ~\citep{JMLR:v15:srivastava14a} that randomly breaks up these complex relationships, Neuron-specific dropout helps networks to understand which of these complex relationships are causing it to overfit, and shuts off hidden units to force the network to learn without them. Using NSDropout proved to improve the performance of neural nets in image classification domains. NSDropout was able to achieve best-in-class results in MNIST Handwritten Digits, Fashion-MNIST, and CIFAR-10. 

In addition, to improve the results of image classification networks, NSDropout also reduces the need for large data sets. When training on MNIST Handwritten Digits, a NSDropout network was able to achieve a perfect test accuracy using only 750 training examples. In Fashion-MNIST, NSDropout was able to achieve a near-perfect accuracy using only 10,000 of the available 60,000 training examples. One key characteristic of NSDropout is its ability to tie the testing accuracy to the training accuracy during training. This helps to limit the chance of a network overfitting. 

One limitation to NSDropout is the increase in the time it takes to train a model. An image classification NSDropout model can take up to 4 times longer to train than a standard neural network of the same architecture with no optimizations made. It can take up to 2 times longer than a traditional dropout model. One major cause of the increase in time is the arrangement and disarrangement of multiple inputs by class into the NSDropout layer. While sorting algorithms have gotten faster, and more optimizations could be done to NSDropout, they still take up the majority of the time during processing. For now NSDropout just ``drops" a unit it finds a network to be too reliant on, but future work could look into how adjusting the unit instead of dropping it could improve the performance in a wider array of applications.

\appendix
\section*{Appendix A.}

Training a neural network requires a vast amount of hyperparameter tuning. NSDropout networks do not help in this regard and even add a hyperparameter. This section describes the heuristics that are relevant for training a model using NSDropout. Most, but not all, heuristic recommendations described in dropout ~\citep{JMLR:v15:srivastava14a} are relevant for NSDropout.

\subsection*{A.1 Network Size}

Similar to dropout, NSDropout decreases the expected capacity of a neural network. Assuming a network has $n$ hidden units in any layer, and $p$ is the proportion of hidden units retrained, the true capacity of the network is $np$ for each layer. The values retrained after NSDropout will follow a similar curve to Figure 9. As a result, the expected change in behavior of a network after training would be similar to a network with $np$ hidden units. Therefore, if an ideal network has $n$ hidden units, an ideal NSDropout network will have $n/p$ units. 

\subsection*{A.2 Learning Rate and Momentum}

Like any network, assigning a learning rate too small, even with NSDropout, will cause it to be stuck in a local minimum. When using optimal learning rates for standard neural networks, NSDropout is still able to achieve high accuracy, but increasing the learning rate by 10-50 times the original is a fairly safe way to speed up training. NSDropout has been shown to have an extensive effect on how a network performs so depending on the task the optimal learning rate might need to be decreased or increased from the above-recommended values. The same can be said for momentum. While a momentum value of 0.9 is common for a standard net, NSDropout might improve with a value closer to 1. As with the learning rate, the momentum of a network might need to be adjusted above or below the recommended.

\subsection*{A.3 Dropout Rate}

When training a network with NSDropout the proportion of units retained, $1-p$, introduces a new hyperparameter. A $p$ value of 0.2 implies that 80\% of units will be retained. A $p$ value of 0 implies that all units will be retained. For image classification it is recommended that 0.5 $\leq p \leq$ 0.7. For the hidden layer, it is recommended that 0.2 $\leq p \leq$ 0.4 but like any hyperparameter, it is recommended to adjust the value based on validation data.

\section*{Appendix B. Detailed Description of Experiments and Data Sets}

Network architectures and training details for the experimental results reported in this paper can be found in this section. The code for reproducing these results can be found at \url{https://github.com/JoshuaShunk/NSDropout}. No machine learning libraries were used and all code is in raw python and NumPy thanks to \cite{kinsley_kukiela_2020}.

\subsection*{B.1 MNIST Handwritten Digits}

The MNIST Handwritten Digits data set contains 60,000 training and 10,000 test examples from 10 different classes (digits 0-9). When testing traditional and dropout networks 20\% or 12000 of the 60000 training images were set aside for validation. When using NSDropout, it was only necessary to use a max of 10000 images. Thus, 2000 of the images were set aside for validation and any given samples of the remaining 50000 images could be used for unseen validation. Hyperparameters were tuned based on both the validation and unseen validation data. The trend from one epoch to the next is more unpredictable than a traditional network as seen in Figure 5. As a result, traditional, dropout, and NSDropout models, were trained for 100 epochs, and were then retrained, stopping at the best epoch found when training prior. 

The architecture shown in Figure 5 was used for all testing of MNIST Handwritten Digits. Each layer had between 128 and 256 units. The architecture using 256 units was most commonly used for traditional networks as 128 units were found to not be enough to provide competitive results. Network with more than 1024 units per layer took too long to train to be analyzed in depth. This is something that will be explored more when NSDropout is optimized for CUDA libraries such as cudamat ~\citep{Mnih09cudamat:a} and cuda-convnet ~\citep{cuda-convnet}.

\subsection*{B.2 Fashion-MNIST}

Inspired by MNIST, Fashion-MNIST was created by~\cite{xiao2017fashionmnist} with the goal of creating a data set with the same level of accessibility as the original MNSIT data set but which was harder in nature. Fashion-MNIST has the same number of samples, 60,000 training, and 10,000 testing, as the original MNIST with the same size examples, 28 x 28, as well as the same number of image categories. The same process that was used for MNIST was used for Fashion-MNIST. When training traditional and dropout networks, 12,000 or 20\% of images were set aside for validation while the rest of the 48,000 were used for training. When training a NSDroput network a max of 10,000 images was necessary for training. Of the 10,000 training images, 2,000 were set aside for validation when the other 8,000 were used for training. Any of the remaining 50,000 could be used for unseen validation. The same idea of early stopping was used for Fashion-MNIST. 

As a result of the hard nature of Fashion-MNIST, the only difference between MNIST and Fashion-MNIST architecture was a change from 128 units per layer to 256. Testing on standard neural networks remained 256 units per layer.

\subsection*{B.3 CIFAR-10}

The CIFAR-10 data set consists of 50,000 training and 10,000 test images. Images are each in one of 10 classes and are represented by a 32 x 32 color image. When training traditional and dropout networks 10,000 of the 50,000 images were set aside for validation while the remaining 40,000 were used for training. When training NSDropout only 25,000 of the original 50,000 images were necessary for training. The original (50000,32,32,3) training size when then resized to (50000,3072) before training. Each network used 256 units per layer for training. NSDropout was applied to the input layer with a $p$ value of 0.5 was used alongside NSDropout was applied to the first hidden layer with a $p$ value of 0.2. Preprocessing using ZCA whitening was also attempted. Develop by~\cite{BELL19973327}, ZCA is an image preprocessing method with the goal of making lines in an image more prominent. Results of applying ZCA to CIFAR-10 images are shown in Figure 6a. Using ZCA showed no noticeable improvement to the results of NSDropout.

\vskip 0.2in
\newpage

\bibliography{main}

\end{document}